\pdfoutput=1

\documentclass[11pt]{article}

\usepackage[final]{acl}

\usepackage{times}
\usepackage{latexsym}

\usepackage[T1]{fontenc}

\usepackage[utf8]{inputenc}

\usepackage{microtype}

\usepackage{inconsolata}

\usepackage{graphicx}

\usepackage[labelfont=bf]{caption}

\RequirePackage{color}

\usepackage{xspace}
\newcommand{\pixel}[0]{\textsc{pixel}\xspace}
\newcommand{\bert}[0]{\textsc{bert}\xspace}
\newcommand{\vitmae}[0]{\textsc{ViT-mae}\xspace}

\let\task\texttt

\newcounter{researchquestion}
\newcommand{\researchquestion}[1]{\refstepcounter{researchquestion}\label{#1}\paragraph{RQ\theresearchquestion:}}

\makeatletter
\def\sectionautorefname{\S\@gobble}
\def\subsectionautorefname{\S\@gobble}
\def\subsubsectionautorefname{\S\@gobble}
\makeatother

\usepackage{booktabs, multirow} %
\usepackage{soul}%

\title{Pixology: Probing the Linguistic and Visual Capabilities of \\
Pixel-based Language Models}

\author{Kushal Tatariya$^{1}$ \ \ Vladimir Araujo$^{2}$ \ \ Thomas Bauwens$^{1}$  \ \ Miryam de Lhoneux$^{1}$ \\
         $^1$ Department of Computer Science, KU Leuven  \\
          $^2$ Sailplane AI \\
           \texttt{kushaljayesh.tatariya@kuleuven.be}\\ 
           }

\begin{document}
\maketitle
\begin{abstract}
Pixel-based language models have emerged as a compelling alternative to subword-based language modelling, particularly because they can represent virtually any script. \pixel, a canonical example of such a model, is a vision transformer 
that has been pre-trained on rendered text. While \pixel has shown promising cross-script transfer abilities and robustness to orthographic perturbations, it falls short of outperforming monolingual subword counterparts like \bert in most other contexts. This discrepancy raises questions about the amount of linguistic knowledge learnt by these models and whether their performance in language tasks stems more from their visual capabilities than their linguistic ones. To explore this, we probe \pixel using a variety of linguistic and visual tasks to assess its position on the vision-to-language spectrum. Our findings reveal a substantial gap between the model’s visual and linguistic understanding. The lower layers of \pixel predominantly capture superficial visual features, whereas the higher layers gradually learn more syntactic and semantic abstractions. Additionally, we examine variants of \pixel trained with different text rendering strategies, discovering that introducing certain orthographic constraints at the input level can facilitate earlier learning of surface-level features. With this study, we hope to provide insights that aid the further development of pixel-based language models.
\footnote{\href{https://github.com/kushaltatariya/Pixology.git}{\texttt{https://github.com/kushaltatariya/Pixology}}}
\end{abstract}

\section{Introduction}
Subwords are currently the standard units of processing in language modelling \cite{sennrich_neural_2016}. While they have been shown to work well in monolingual models \citep{devlin_bert_2019, roberta_paper}, in a multilingual context they can lead to an inevitable vocabulary bottleneck with each language competing for space in a finite vocabulary \cite{rust2023language,liang-etal-2023-xlm}. Characters and byte-based models have been proposed as alternatives to subwords, but they lead to longer input sequences \cite{JMLR:v21:20-074,xue-etal-2022-byt5,tay2022charformer,10.1162/tacl_a_00448}. Another proposed solution is pixel-based models where patches of pixels are the main unit of representation. A canonical example of this is the \pixel (Pixel-based Encoder of Language) model \cite{rust2023language}, where text is rendered as a sequence of fixed-sized patches and passed as input to a vision transformer (ViT) \citep{dosovitskiy2021an}. This approach allows the model to represent virtually any script.

Although current versions of the pixel-based language models do not outperform their monolingual subword-based counterparts on most downstream tasks \citep{rust2023language,lotz-etal-2023-text}, they are a promising approach to multilingual modelling and offer a unique opportunity to explore modelling language through images. \pixel is a juxtaposition of a vision and language model: even though it receives image patches as input, the content of those patches is rendered text, making it a visual model of language. With this study, we aim to understand where \pixel stands on the vision-to-language spectrum. To this end, we probe \pixel on various visual and language tasks and compare performance with \bert \cite{devlin_bert_2019} -- the language model it is most comparable to -- and \vitmae \cite{9879206} -- the vision model it is most comparable to. We conduct a comprehensive analysis of the linguistic and visual capabilities of \pixel that can be used to aid further development of pixel-based language models. Concretely: 
\researchquestion{rq:linguistic} How much linguistic knowledge is encoded in \pixel?
\researchquestion{rq:visual} How much visual capability does \pixel have?\footnotemark

We find that \pixel learns surface-level linguistic information in the lower layers, resulting in higher-level syntactic and semantic abstractions appearing in higher layers than \bert (\autoref{sec:rq1}). When comparing to \vitmae, \pixel underperforms on image tasks, with visual probing accuracy decreasing in the higher layers (\autoref{sec:rq2}). Thus, the surface-level information is diluted as it acquires linguistic knowledge in the higher layers. 

\citet{lotz-etal-2023-text} trained newer pixel-based language models that add some orthographic constraints to the input that can potentially augment linguistic learning in the lower layers. In this context, we ask the following question:

\researchquestion{rq:rendering} Does adding orthographic constraints to the input enhance the linguistic capabilities in \pixel?

\footnotetext{By visual capability, we refer to a surface level understanding of characters in a text, analogous to the kind described in \cite{conneau-etal-2018-cram}.}

We find that a rendering strategy that makes word boundaries more explicit in the input enables \pixel to learn surface-level linguistic features earlier in the model, thereby aiding semantic understanding (\autoref{rq3}). 

Overall, we take inspiration from BERTology, the study of the 
linguistic capabilities in \bert \cite{rogers-etal-2020-primer}, and aim for this work to foster future explorations and advancements for \pixel.

\section{Background}
\subsection{PIXEL}
The pixel-based language models examined in this study are ViTs \cite{dosovitskiy2021an} that lie at the confluence of NLP and computer vision. A ViT is an application of the transformer architecture \citep{vaswani_attention_2017, devlin_bert_2019} to process images. 
An image is split into patches that are each flattened into a vector and then projected into a lower-dimensional space through a linear transformation. Positional embeddings are added to retain spatial information 
before feeding these patch vectors into a transformer encoder. 

Inspired by the self-supervised masked language modelling paradigm, a variant of ViT is the masked auto-encoder \cite{9879206}, or \vitmae, that learns image representations by masking random image patches. A decoder
reconstructs the image from the latent representation of the mask tokens. The \pixel model by \citet{rust2023language} is trained on the \vitmae architecture. 
It takes a rendered image of text sized $16 \times 8464$ as input, which is split into patches of $16\times 16$ pixels. Instead of randomly masking \emph{individual} patches, \pixel randomly masks \emph{spans} of patches to force the model to learn higher levels of language abstraction.
\pixel is pre-trained on a rendered version of the English Wikipedia and the BookCorpus \cite{Zhu_2015_ICCV}. Thus, it is comparable to \bert in terms of pre-training data and \vitmae in terms of architecture and parameters.

\pixel follows the idea of visual text representations by \citet{salesky-etal-2021-robust}, who embed rendered text using 2D convolutions for continuous open-vocabulary machine translation. They demonstrate that visual text representations are more robust to noise and provide the benefits of a tokenization-free text processing pipeline. 
In other applications, \citet{borenstein-etal-2023-phd} examined the benefits of using pixel-based encoders for an OCR-free approach to language modelling of historical documents. 

\citet{lotz-etal-2023-text} further improved \pixel by experimenting with different text rendering strategies. 
Their work provides insights into the semantic modelling capabilities of \pixel models and correlates that to frequency bias. We include some of these models in our study.

\begin{table*}[!htp]\centering
    \captionsetup{font=footnotesize}
    \footnotesize
    \begin{tabular}{lrp{0.4\linewidth}r}\toprule
        \textbf{Type} &\textbf{Name} &\textbf{Predict for a given sentence...} &\textbf{Labels} \\\midrule
        \multicolumn{4}{c}{\textbf{Linguistic Probing}} \\\cmidrule{1-4}
        \multirow{2}{*}{\textbf{Surface}} &Sentence Length (\textbf{SentLen}) &the length. &6 bins \\
        &Word Content (\textbf{WC}) &which one of 1000 possible words is in it. &1000 \\\midrule
        \multirow{3}{*}{\textbf{Syntactic}} &Bigram Shift (\textbf{BShift}) &if the order of two random words was inverted. &2 \\
        &Tree Depth (\textbf{TreeDepth}) &the depth of the syntactic tree. &5-12 \\
        &Top Constituents (\textbf{TopConst}) &the sequence of top constituents directly below the sentence (S) node. &20 \\\midrule
        \multirow{3}{*}{\textbf{Surface Semantic}} & Tense (\textbf{Tense}) &the tense of the main verb. &3 \\
        &Subject Number (\textbf{SubjNum}) &the number of the subject. &2 \\
        &Object Number (\textbf{ObjNum}) &the number of the object. &2 \\\midrule
        \multirow{2}{*}{\textbf{Complex Semantic}} &Semantic Odd Man Out (\textbf{SOMO}) &if a noun or a verb has been switched out for another. &2 \\
        &Coordination Inversion (\textbf{CoordInv}) &if the two coordinate clauses have been inverted. &2 \\\midrule
        \multicolumn{4}{c}{\textbf{Visual Probing}} \\\midrule
        \multirow{2}{*}{\textbf{Visual}} 
        &Max Count (\textbf{MaxCount}) &the frequency of the character with the max count. &4 bins \\
        &Argmax Count (\textbf{ArgmaxCount}) &the character that has the max count. &5 bins \\
        \bottomrule
    \end{tabular}
    \caption{Description of probing tasks used in this study.}\label{tab:SentEval Tasks}
\end{table*}

\subsection{Model Interpretability}
The survey by \citet{10.1145/3639372} categorises model interpretability into local explanations of predictions and global explanations of model behaviour. 
Global explanations 
aim to understand the general concepts encoded in the individual components of a language model. 
The most prominent method for global explanations of linguistic understanding in language models is \emph{probing}, specifically classifier-based probing \cite{belinkov-2022-probing}. 

In this approach, model weights are frozen and for each of its layers, a small classifier is trained to solve a task given a pooled representation of the intermediate embeddings at that layer. The task is designed to isolate an aspect of linguistic understanding that may or may not be present in the embedding \cite{DBLP:journals/corr/AdiKBLG16, hewitt-manning-2019-structural, sahin-etal-2020-linspector, zhu-etal-2022-predicting}.
The same idea has been used for investigating computer vision models \cite{alain2018understanding,ijcai2021p82}
and, more recently, multi-modal models \cite{dahlgren-lindstrom-etal-2020-probing}.

A standard framework for linguistic probing is SentEval \cite{conneau-kiela-2018-senteval}, which includes various probing tasks that uncover different levels of linguistic information in sentence embeddings. SentEval has been extensively employed to analyse models for sentence-level semantics \cite{ma_universal_2019, krasnowska-kieras_empirical_2019, ravichander_probing_2021}, and it is the dataset we adopt in this study. 

Linguistic probing has been used prominently in BERTology \cite{rogers-etal-2020-primer} to understand the levels of linguistic information stored in \bert embeddings \cite{tenney2018what, jawahar-etal-2019-bert, mehrafarin-etal-2022-importance}. 
It has been established that \bert tends to encapsulate more syntactic knowledge in its middle layers, while semantic comprehension is more pronounced in the higher layers \cite{tenney-etal-2019-bert}. In this context, we aim to gain analogous insights into pixel-based language models.

\section{Probing Tasks}
We now introduce the probing tasks used in our experiments. We probe \pixel on two levels: linguistic and visual. For linguistic probing, we rely on the SentEval framework. 
ViTs have more direct access to surface-level information than subword-based models since their input is segmented into units of fixed visual size (as opposed to variable-sized tokens) and shown to the model after a continuous linear projection (as opposed to a lookup). Thus, we also employ tasks that are designed to verify whether orthographic information is more easily identifiable throughout \pixel.

\subsection{Linguistic Probing Tasks}
The SentEval framework contains probes that quantify three levels of linguistic knowledge present in sentence embeddings: \emph{surface}, \emph{syntactic}, and \emph{semantic} \citep{conneau-etal-2018-cram}. \autoref{tab:SentEval Tasks} presents all the tasks, with their type and description. We evaluate the performance of the models at each layer on these tasks to explain the hierarchy of linguistic understanding contained within the model. 

We note, however, that all tasks falling under the \emph{semantic} category do not all probe for the same kind of information.  \task{Tense}, \task{SubjNum} and \task{ObjNum} can be solved by trivial surface cues like the presence of certain morphemes like the suffixes \textsl{-ed} and \textsl{-es}. However, unlike surface tasks, performance on these tasks does not drastically degrade in the upper layers as the model gains semantic understanding \cite{jawahar-etal-2019-bert}, and they can be predictors of downstream semantic performance \cite{zhu-etal-2022-predicting}. Thus, we dub these tasks \emph{surface semantic}. 

\task{SOMO} and \task{CoordInv}, on the other hand, need more complex semantic learning to be solved. We therefore term these tasks as \emph{complex semantic}. The distinction between \emph{surface semantic} and \emph{complex semantic} can also be justified by the differences in accuracy between human evaluation and model performance for these tasks as reported by \citet{conneau-etal-2018-cram}. Most neural models are able to either match or surpass human evaluation for the \emph{surface semantic} tasks, but not for the \emph{complex semantic} tasks. This re-categorization also helps to identify consistencies in linguistic understanding, particularly when explaining trends with \bert.

\subsection{Visual Probing Tasks}\label{subsec:visual-probing}

We introduce two new tasks to probe for purely visual information -- \task{MaxCount} and \task{ArgmaxCount} (see \autoref{tab:SentEval Tasks}). Every word in every sentence of the \task{SentLen} task is replaced by a random English word generated with the \texttt{wonderwords}\footnote{\href{https://github.com/mrmaxguns/wonderwordsmodule}{\texttt{github.com/mrmaxguns/wonderwordsmodule}}} 
library to create synthetic datasets. By using random words instead of a sentence, we ensure that the task is purely visual, but does not disadvantage the \bert tokenizer (as opposed to using random characters which could result in single-character tokens). This also distinguishes them from the \emph{surface} tasks in SentEval since there is no underlying linguistic pattern to this data. The labels are binned to ensure a uniform distribution and we down-sample the labels that occur with a very high frequency (for example, `e' is the most frequent letter in 50\% of the dataset). More task details are in \autoref{sec:visual-app}.

\paragraph{MNIST}
As a final task to probe for purely visual information, we rely on MNIST \cite{6296535}, which consists of white-on-black images of handwritten digits (0 to 9). It is an image classification benchmark dataset and its resemblance to rendered text as well as the simplicity of the task make it suitable for probing.\footnote{Each image is $28 \times 28$ pixels that we resize to $16 \times 16$, the image patch size for one patch in \pixel.}
We do not evaluate \bert on this task since it cannot represent images.

\section{Experimental Setup}
\subsection{Models}\label{sec:models}
Our analysis will primarily focus on the \pixel-base model trained by \citet{rust2023language}, further termed \pixel. We also make a comparison with its variants introduced by \citet{lotz-etal-2023-text} for \textbf{RQ3}. Specifically, we look at \pixel-bigrams, pre-trained using the \texttt{bigrams} rendering strategy which ensures that every patch contains at most 2 characters, and that no patch overlaps a word boundary, adding extra space where needed.
We also look at \pixel-small-words, trained on the \texttt{words} rendering strategy that merely enforces the second constraint. Since it has no base version released, we additionally probe \pixel-small-bigrams and \pixel-small for a fair comparison.\footnote{\href{https://huggingface.co/Team-PIXEL}{\texttt{huggingface.co/Team-PIXEL}}} All these are compared against \bert and \vitmae. An overview of the model parameters is in \autoref{sec:model_params}.

\subsection{Probing}
We follow the same probing setup as defined by \citet{conneau-kiela-2018-senteval}. Sentence representations for each example in the datasets are obtained by mean-pooling the token or patch embeddings generated at every hidden layer for each model. 
These embeddings are passed to a classifier that learns to predict the corresponding class label using a cross-entropy loss. For our experiments, we use the implementation and default hyper-parameters proposed by \citet{araujo-etal-2022-evaluation} for both linguistic and visual tasks.

\subsection{Fine-tuning}
For a better understanding of the general linguistic abilities of vision models (\textbf{RQ1}), we fine-tune \vitmae on universal dependencies (UD) \cite{nivre-etal-2016-universal} POS-tagging, dependency parsing and GLUE \cite{wang-etal-2018-glue} using the same hyperparameters as \citet{rust2023language}. We re-use \pixel's text rendering configuration, and render text into a square image of $224 \times 224$ to match the input size of \vitmae. To gauge the general visual abilities of \pixel (\textbf{RQ2}), we fine-tune \pixel and \vitmae on the CIFAR100 \citep{Krizhevsky09} image classification dataset.

\section{Results and Analysis}

\subsection{RQ1: How much linguistic knowledge is encoded in \pixel?} \label{sec:rq1}

To investigate this question, we first compare \pixel and \vitmae fine-tuned on language tasks. This is to assess the extent to which \pixel's pre-training regime makes it better at language tasks than a regular vision transformer.
Results are in \autoref{tab:GLUE}.

\begin{table}[!htp]\centering
    \captionsetup{font=footnotesize}
    \footnotesize
    \begin{tabular}{lrrrr}\toprule
        \textbf{Task} &\textbf{\pixel} &\textbf{\vitmae} &\textbf{\bert} \\\midrule
        PoS &0.97 &0.93 &0.97 \\
        DP &0.89 &0.68 &0.91 \\
        GLUE avg. &0.74 &0.58 &0.80 \\
        \bottomrule
        \end{tabular}
    \caption{Language fine-tuning accuracy for \pixel, \bert (taken from \citet{rust2023language}) and \vitmae}\label{tab:GLUE}
\end{table}

\begin{figure*}
    \captionsetup{font=footnotesize}
    \centering
    \scriptsize
    \includegraphics[scale=0.77]{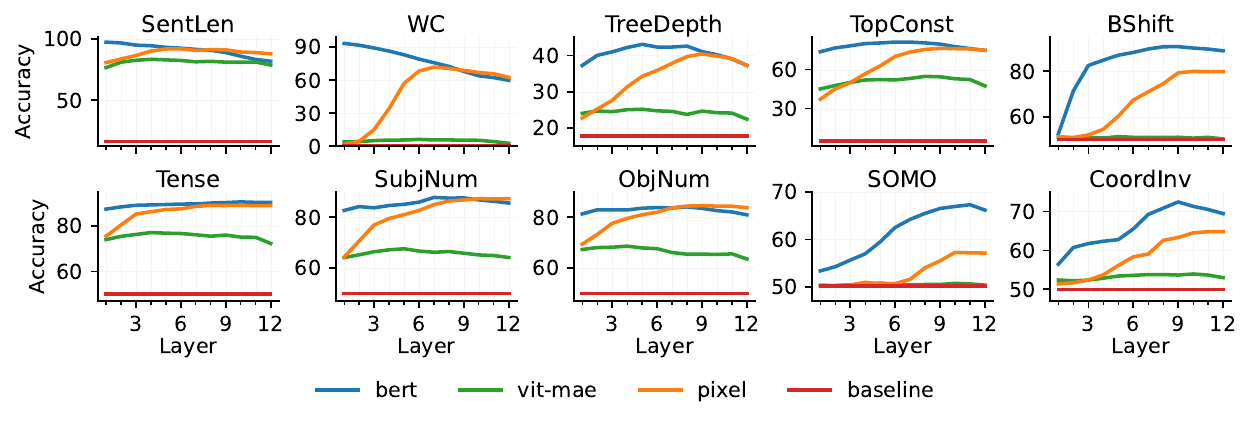}
    
    \caption{Linguistic probing results for layers 1-12 of \pixel, \bert and \vitmae, along with the majority baseline.}
    \label{fig:senteval-base-probe}
\end{figure*}
It is clear that \pixel has an advantage over \vitmae. Since \pixel performs substantially better than \vitmae on GLUE, it can be argued that \pixel learns some semantics. This can be explained either by the domain similarity between \pixel pre-training and the downstream task input 
or because its pre-training on language actually enables the model to learn linguistic abstractions.

\begin{figure} 
    \captionsetup{font=footnotesize}
    \centering
    \includegraphics[scale=0.07]{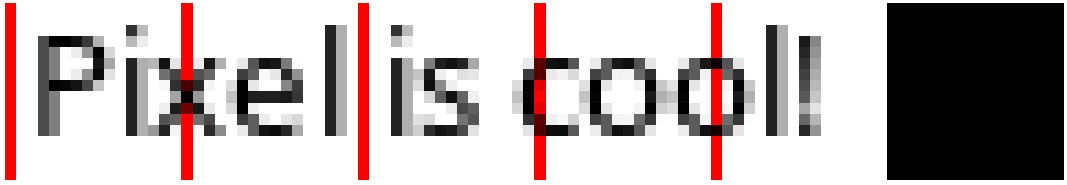}
    \includegraphics[scale=0.07]{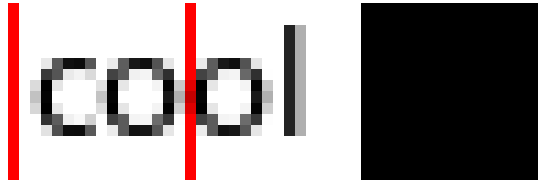}
    \caption{Example of "cool" being rendered differently in different contexts for \pixel. The red lines represent patch boundaries.}
    \label{fig:WORD_EMB}
\end{figure}

To investigate which of the two factors explains the advantage, we run the linguistic probing tasks on \pixel, \bert, and \vitmae, illustrated in \autoref{fig:senteval-base-probe}. Each plot also includes the majority baseline\footnote{The accuracy if always predicting the most frequent label.} for that task as a lower bound for each model. If the embeddings do not contain any useful information for the task, we would expect the performance to be equivalent to the majority baseline.

The performance of \bert is consistent with what is documented in literature. Surface features are encoded in the lower layers, syntactic features are represented in the middle layers, and semantic features are found in the upper layers \cite{jawahar-etal-2019-bert}. The performance for \vitmae for all layers, for most tasks, is very close to the majority baseline. For tasks where some visual information can be useful, for example in \texttt{SentLen}, and \texttt{Tense} (the visual presence of morpheme \textsl{-ed} can be associated with label \texttt{PAST}), \vitmae performs better than the majority baseline but does not improve or decline through the layers. The performance of \pixel, when higher than \vitmae, can thus be attributed to its linguistic knowledge and not due to having input that is closer to the downstream task.

Across all tasks, \pixel consistently has an initial monotonic rise in accuracy, starting with a similar performance as \vitmae in the lower layers. This indicates that it is using purely visual information in the lower layers, and learns linguistic information in the higher layers. In other words, \pixel starts as a visual model, and becomes more of a language model through the layers.  

However, \pixel never matches the peak performance of \bert in any layer. This is consistent with the results from \citet{rust2023language}, where \pixel underperforms \bert on both syntactic and semantic downstream English tasks. We can, therefore, hypothesize that much of \pixel's capacity is used in recovering from the performance gap between a vision and language model. 
 
\paragraph{Does \pixel learn syntax and semantics?}
Unlike \bert, \pixel does not have a consistent curve across the \emph{surface}, \emph{syntactic} and \emph{semantic} tasks. This is most striking in the \emph{surface} tasks. For \bert, there is an inverse relation between model depth and accuracy. For \texttt{SentLen}, the accuracy curve of \pixel rises until layer 5 and then stagnates. For \texttt{WC}, on the other hand, it has a steep rise in the initial layers until layer 7, where it drops. 
The task requires a good understanding of word-level features and boundaries -- something that is encoded in \bert already at the input level, but \pixel has to learn during training. We illustrate this further in \autoref{fig:WORD_EMB}. 
The patches encoding the word "cool" differ when used in the context of a sentence compared to when it is rendered alone 
Thus, it may take more layers for \pixel to reconcile the two different embeddings as the same word. \citet{lotz-etal-2023-text} have also commented on this phenomenon and linked it to poor downstream semantic performance. They also found that \pixel-based language models form better contextualised word representations in the upper layers of the model.

We can extrapolate this phenomenon to explain the initial monotonic rise in other tasks. For \emph{syntactic} tasks, \pixel peaks at layer 9, later than \bert, then stagnates or declines. This leads to a delayed learning of higher level abstractions, suggesting that \pixel needs more layers to match \bert's performance. We leave this exploration to future work.

The performance across the \emph{surface semantic} tasks for \pixel shows some consistency. There is a steep rise until layer 3, after which the curve has a more gradual rise, crossing \bert accuracy in the higher layers. For \emph{complex semantic} tasks, both \pixel and \bert achieve peak performance between layers 9 and 12. However, the performance gap between the two is substantial, indicating that \pixel does not learn semantic abstractions at the same level as \bert.
This is also substantiated by the difference in the downstream performance gap between \bert and \pixel for syntactic and semantic tasks, mentioned in \autoref{tab:GLUE}. \pixel's performance on dependency parsing and POS-tagging is very close to \bert, while its performance on GLUE, which contains tasks requiring more semantic understanding, is about 6\% lower.

The drop in performance for surface tasks in the higher layers also indicates that \pixel forgets some surface level information as it learns more linguistic abstractions. We substantiate this further with the results on the visual probing tasks below.

\subsection{RQ2: How much visual capability does \pixel have?}
\label{sec:rq2} 

\begin{figure}
    \captionsetup{font=footnotesize}
    \centering
    \includegraphics[scale=0.75]{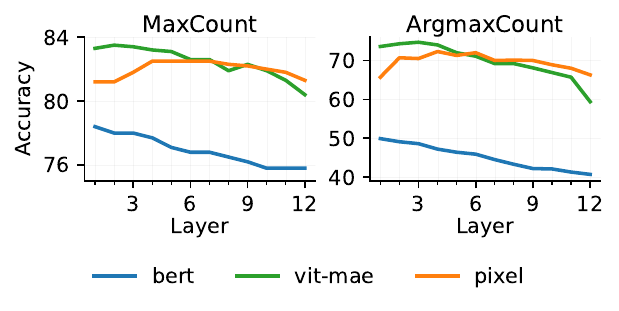}

    \caption{Visual probing results for layers 1-12 of \pixel, \vitmae and \bert.}
    \label{fig:visual_probing}
\end{figure}

We investigate this question by first probing \pixel on the visual tasks introduced in \autoref{subsec:visual-probing} to understand whether it is indeed forgetting the surface level information in the higher layers. Results are shown in \autoref{fig:visual_probing}.

For both \texttt{MaxCount} and \texttt{ArgmaxCount}, we see that \vitmae has the highest performance in the lower layers, followed by \pixel and then \bert. \bert performance has a steady decline, much like the \emph{surface} tasks in \autoref{fig:senteval-base-probe}. \pixel's performance is much closer to \vitmae, but it does not have much decline through the layers, leading to a higher performance than \vitmae in the higher layers. \pixel has slight increases in performance in the middle layers, analogous to the performance peaks in \emph{surface} tasks. The substantially higher performance than \bert, combined with the similarity to \vitmae performance, indicates that \pixel still retains much surface level information in the higher layers. \pixel's high performance on \textit{surface semantic} tasks in the higher layers also substantiate this since \pixel has access to both surface and semantic information. 

\paragraph{Can \pixel be a vision model?}
If \pixel still retains much surface information in the higher layers, is it able to perform well on vision tasks? To investigate this question, we present fine-tuning results for \pixel and \vitmae in \autoref{tab:cifar100}. If \pixel performs competitively, it implies that \pixel is fundamentally a vision model that has acquired some language understanding. We also fine-tune a transformer of the same size with randomized weights as a lower bound baseline.

\begin{table}[!htb]\centering
    \captionsetup{font=footnotesize}
    \footnotesize
    \begin{tabular}{lrr}\toprule
        \textbf{Model} &\textbf{Accuracy} \\\midrule
        \pixel &0.52 \\
        \vitmae &0.83 \\
        Random Model &0.42 \\
        \bottomrule
    \end{tabular}
    \caption{Results for \pixel, \vitmae and \vitmae with randomised weights fine-tuned on CIFAR100 for image classification.}\label{tab:cifar100}
\end{table}
The performance gap between \pixel and \vitmae on image classification is analogous to the performance gap between the two on the GLUE tasks in \autoref{tab:GLUE}. Thus, even though \pixel is a vision transformer and it retains much surface level information, its pre-training regime on language has lead to a substantially worse performance on image classification, much closer to the random baseline than to \vitmae. 
It can be argued that \pixel's poorer performance on CIFAR-100 is due to a domain mismatch, stemming from its pre-training on black-and-white text, which offers limited exposure to the color and complexity of the input.

\begin{figure}
    \captionsetup{font=footnotesize}
    \centering
    \includegraphics[scale=0.75]{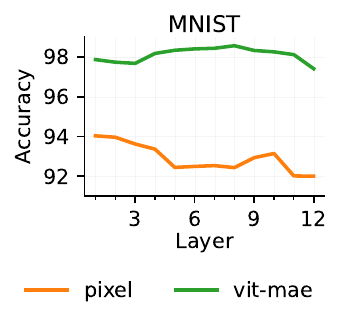}
    \caption{MNIST probing results for layers 1-12 of \pixel and \vitmae.}
    \label{fig:MNIST}
\end{figure}

To disentangle this, we probe \pixel on MNIST at every layer. The results are in \autoref{fig:MNIST}. The curves for \pixel are consistent with the curves in \autoref{fig:visual_probing} and \textit{surface} tasks in \autoref{fig:senteval-base-probe}, in that there is a performance decline through the layers. The difference is that \pixel's performance declines immediately after layer 1, and unlike \autoref{fig:senteval-base-probe}, it is at a lower accuracy than \vitmae in the lower layers. Thus, even on input that is similar to the data that \pixel was pre-trained on, \pixel does not match \vitmae performance.

\subsection{RQ3: Does adding orthographic constraints to the input enhance the linguistic capabilities in \pixel?}\label{rq3}

\begin{figure}
    \captionsetup{font=footnotesize}
    \centering
    \scriptsize
    \includegraphics[scale=0.75]{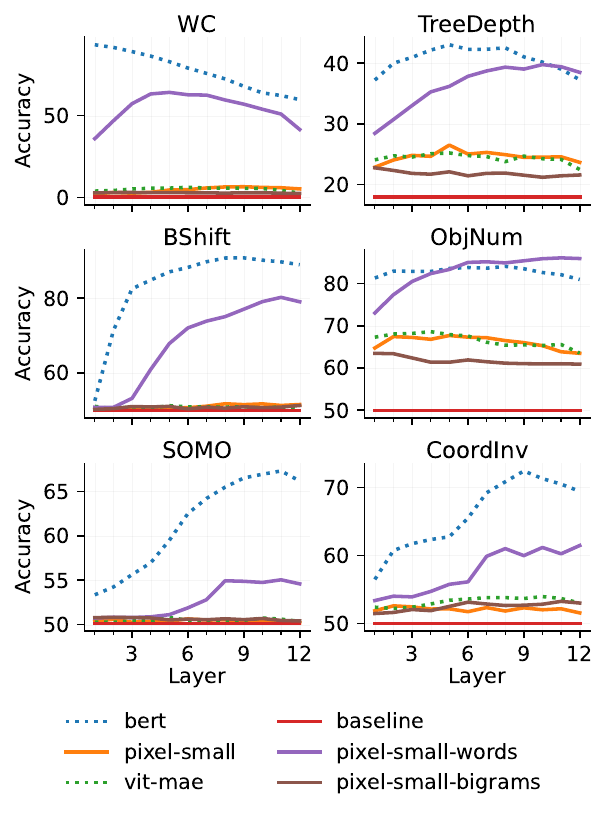}
    
    \caption{Selected linguistic probing results for layers 1-12 of small \pixel variants. Base models are indicated with dotted lines.}
    \label{fig:small-models}
\end{figure}

Results from \S\ref{sec:rq1} and \S\ref{sec:rq2} establish that \pixel learns surface level information in the lower layers, which leads to delayed learning of higher level semantics. This raises the question of how the gap between visual and linguistic understanding in layers 1 - 6 (the layer with peak performance on surface tasks) can be bridged earlier in the model. 
Encoding words with differing visual patch representations, as shown in \autoref{fig:WORD_EMB}, can be made easier by ensuring consistent rendering of words across contexts.
The added constraints to the rendering in the \pixel-variants may lead to a faster learning of surface level information and word boundaries in the lower layers, as discussed in \S\ref{sec:rq1}, thereby making 
\pixel behave more like \bert. This idea is further substantiated by the fact that \pixel-bigrams and \pixel-small-words have better downstream performance than \pixel. Probing results on selected tasks for \pixel-small, \pixel-small-words, and \pixel-small-bigrams are in \autoref{fig:small-models}. We also include \bert-base and \vitmae-base in the graphs for reference.

At the small scale, \pixel suffers an almost catastrophic decline in performance, showing no more linguistic understanding than \vitmae. Similarly, \pixel-small-bigrams also does not demonstrate any meaningful linguistic understanding. \pixel-small-words, on the other hand, displays probing performance comparable to \pixel-base, even at the small scale. 
It starts with much higher accuracy than \vitmae in layer 1 -- indicating that there is already linguistic information present in the initial layers due to the imposed structure at the input level. It also achieves peak performance in most tasks earlier than \pixel-base\footnote{We speculate that \pixel-small-words outperforms \pixel-small-bigrams, even though both prevent patch overlap across word boundaries, because the extra space added by \texttt{bigrams} rendering to maintain two characters per patch leads to a loss of word boundary information and longer sequences. A deeper exploration is left for future work.}. 

Specifically, for \texttt{WC}, the accuracy rises only until layer 4 before it declines. The curves for \textit{syntactic} tasks are more similar to \bert, with the lower layers achieving scores higher than \pixel. A combination of visual and some semantic understanding leads to scores for \textit{surface semantic} tasks being even higher than \bert in the upper layers. For \textit{complex semantic} tasks, however, the curve rises until layers 7-8 and then plateaus, indicating higher semantic abstractions are still not being learnt by the model.

Since \pixel-small-bigrams and \pixel-small do not have any meaningful linguistic representations at the small scale, we also compare the base version of the two models on the linguistic probes to find similar trends. \pixel-bigrams at both the base and small scale performs worse than \pixel. Specific results and analysis can be found in Appendix \ref{pixel-v-bigrams}.

\begin{figure}
    \captionsetup{font=footnotesize}
    \centering
    \scriptsize
    \includegraphics[scale=0.75]{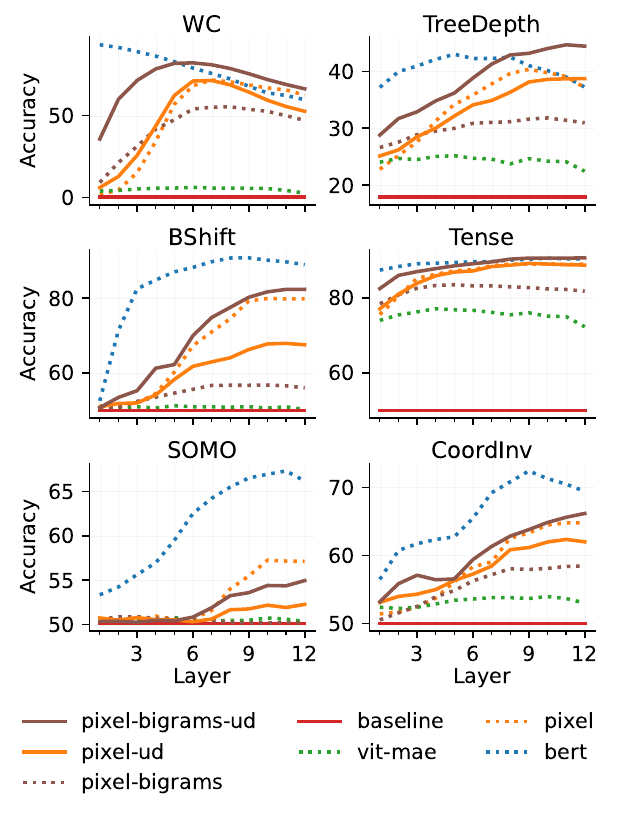}
    
    \caption{Selected probing results for layers 1-12 of \pixel and \pixel-bigrams finetuned on UD.}
    \label{fig:UD-probing}
\end{figure}

\begin{figure}
    \captionsetup{font=footnotesize}
    \centering
    \scriptsize    
    \includegraphics[scale=0.75]{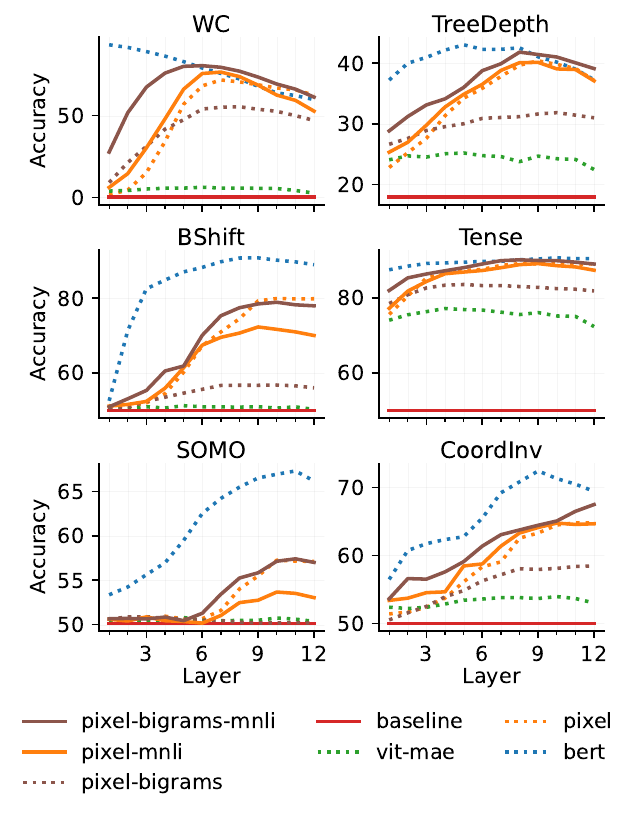}
    
    \caption{Selected probing results for layers 1-12 of \pixel and \pixel-bigrams finetuned on MNLI.}
    \label{fig:mnli-probing}
\end{figure}

\paragraph{Why is fine-tuned \pixel-bigrams better than fine-tuned \pixel?}
The observation above is at odds with the downstream performance of \pixel-bigrams, which \citet{lotz-etal-2023-text} found to be better than \pixel. To understand this discrepancy, we run the linguistic probes on fine-tuned versions of the models. We fine-tuned the \pixel-base-bigrams model on UD parsing (syntactic) and MNLI (semantic) with the same hyper-parameter setup as \pixel, and compare them to the fine-tuned \pixel models made available by \citet{rust2023language} on the same tasks. Results are in \autoref{fig:UD-probing} and \autoref{fig:mnli-probing}.

We see that across all probing tasks, fine-tuned \pixel-bigrams demonstrates better performance than fine-tuned \pixel. 
\citet{merchant-etal-2020-happens} found that finetuning \bert on dependency parsing shows effects throughout the model, but MNLI only affects the top layers. Moreover, fine-tuning can cause the model to potentially forget some linguistic knowledge. \citet{mehrafarin-etal-2022-importance} also echoed that fine-tuning on tasks with larger data sizes (like MNLI) can lead to loss of linguistic information in the pre-trained encodings. 

We see this trend in \pixel, where both UD and MNLI fine-tuning decrease probing performance on \task{BShift} and \task{SOMO}. There is a slight decline in performance on all other probing tasks with UD fine-tuning, but with MNLI fine-tuning, the performance remains similar to pre-trained \pixel. 

We observe the contrary with \pixel-bigrams. Both UD and MNLI fine-tuning have enhanced the linguistic knowledge encoded in all the layers, with probing performance compared to \pixel-bigrams pre-trained being much higher. Additionally, UD fine-tuning particularly increases probing performance on \textit{syntactic} tasks in the top layers, and MNLI fine-tuning increases probing performance on the \textit{complex semantic} tasks in the top layers. 

Thus, we can speculate that the inductive bias learnt during fine-tuning creates better linguistic representations in \pixel-bigrams.

\subsection{Summary and Implications of Findings}

On the spectrum of vision and language, it can be concluded from the results of \textbf{RQ1} and \textbf{RQ2} that \pixel is more of a language model than a vision model. The difference in downstream performance between \pixel and \vitmae is much larger than between \pixel and \bert. Although with a lower accuracy, \pixel's behaviour revealed by probing is more similar to \bert than \vitmae.

However, much of \pixel's linguistic knowledge is surface level. The lower layers in \pixel learn surface level information, as demonstrated by the visual and linguistic probes. The linguistic knowledge acquired in the upper layers demonstrates some syntactic capabilities, but does not capture very strong semantic information. This indicates that adding more layers to the model could allow it to have better semantic representations. Other architectural solutions such as a RoBERTa-like \cite{liu_roberta_2019} approach to \pixel pretraining with longer training and a revisiting of the masking and reconstruction method could also be explored. 

While these are solutions on the architecture side, on the input side from the results of \textbf{RQ3} we can conclude that experimenting with the rendering strategies can be very promising. 
\pixel-words emerges as the current best solution to bridging the gap between visual and language understanding in the lower layers in the model. Nevertheless, it still lacks in semantic understanding, and as \citet{lotz-etal-2023-text} have noted it is not very efficient to train. \pixel-bigrams has worse linguistic probing performance than \pixel, but fine-tuning dramatically improves the linguistic knowledge encoded in its layers which could be due to the inductive bias it learns in the process. \pixel, on the contrary, forgets some linguistic information during fine-tuning.

One can argue that the input patches for \pixel-words and \pixel-bigrams most closely resemble the input of a subword-based model, with tokenized units where a patch/token never crosses word boundaries. Even though the model does not have a fixed vocabulary, structured rendering requires some level of pre-tokenization and awareness of linguistic granularity. This raises questions about whether these structured rendering approaches 
might lead to the same issues and debates that surround traditional subword tokenization when it comes to heuristics used for segmentation \cite{clark_canine_2022}, particularly when applied to languages without conventional word boundaries. \citet{lotz-etal-2023-text} have also noted that while structured rendering strategies can give \pixel an advantage, they can also make it difficult to generalise to other languages. Thus, development of \pixel along these lines must be informed by careful consideration of linguistic diversity and the potential limitations posed by structured rendering, ensuring that solutions are adaptable to languages with varied morphosyntactic structures.

\section{Conclusion}

This study is a first step towards understanding the language modelling capabilities of pixel-based models. Although these models exhibit substantial linguistic understanding, the nature of image-text representations leads to a gap in visual and linguistic understanding. Pixel-based models need to learn the discrete representations that subword-based models already have access to at the input level. Adding orthographic constraints to the input can help bridge this gap, but further architectural modifications could improve these models more, which is a promising direction for future work.

\section{Limitations}
Our main approach to understanding the linguistic information encoded in pixel-based language models is probing. We acknowledge that although this is our primary method of inquiry, it comes with its flaws. \citet{belinkov-glass-2019-analysis} have noted that even though certain information is detected by a probe as being present in the embeddings, it does not necessarily imply that the information is used by the model. They also remark that using a deeper auxiliary classifier for the probe may lead to better results. There are other criticisms of the approach like \citet{hewitt-liang-2019-designing} that question whether the probe uncovers information encoded in the embedding, or just learns the linguistic task itself that it is trained on. \citet{pimentel-etal-2020-information} challenge this and present evidence of the former. \citet{zhu-rudzicz-2020-information} recommend using a control mechanism to select probes, based on discussions about the dichotomy raised above. Thus, although this does not dismiss the validity of our findings, we note that our results and conclusions should be read with these caveats in mind.

\section{Acknowledgements}
The computational resources and services used in this work were provided by the VSC (Flemish Supercomputer Center), funded by the Research Foundation - Flanders (FWO) and the Flemish Government - department EWI (for KT, TB and ML). TB is funded by a Bijzonder Onderzoeksfonds (BOF) internal fund at KU Leuven, namely the C1 project fund with reference C14/23/096.

\bibliography{bib/tacl2021.bib, bib/thesis-thomas.bib, bib/custom.bib}

\newpage

\appendix
\section{Visual Tasks}
\label{sec:visual-app}
\paragraph{Max Count Character (MaxCount)} Every letter from the random words in an example is counted. Per example, for the raw counts $\hat f_\ell$ for each letter, we compute $\max_\ell \hat f_\ell$ and split the results into 4 uniformly occurring contiguous bins. The task is to predict this bin given the sentence. Examples where multiple letters have the same maximal count are excluded to ensure that the probing task can only be solved by noticing one particular character. We exclude examples with less than 3 unique characters. Details about labels and frequency of each bin are in \autoref{tab:Max Count}

\begin{table}[!htbp]
\centering
\captionsetup{font=footnotesize}
\footnotesize
\begin{tabular}{lp{0.5\linewidth}rr}\toprule
\multicolumn{3}{c}{\textbf{MaxCount}} \\\cmidrule{1-3}
\textbf{Bin} &\textbf{Labels in Bin} &\textbf{Bin Size} \\\midrule
1 &[3, 4, 5, 6, 7, 8, 9] &21162 \\\midrule
2 &[10, 11, 12, 13, 14] &21162 \\\midrule
3 &[15, 16, 17, 18, 19] &26597 \\\midrule
4 &[20, 21, 22, 23, 24, 25, 26, 27, 28, 29, 30, 31, 32, 33, 34, 35, 36, 37, 38] &25333 \\\midrule
&\textbf{Total} &94,254 \\
\bottomrule
\end{tabular}
\caption{Bin sizes, labels in bin and total data size for \task{MaxCount}. The labels correspond to the count of the character with the maximum frequency in an example.}\label{tab:Max Count}
\end{table}

\paragraph{Argmax count character (ArgmaxCount)} We count the letters in each example again, but now the task is to predict $\ell^* = \arg\max_\ell \hat f_\ell$ given the example. The same examples are excluded as above (meaning the argmax is unique), and we skip examples where the argmax is not one of the 26 lowercase Latin letters $\{a, b, \dots, z\}$. To mitigate against the strong skew towards higher-frequency letters (e, t, a, ...), letters are grouped into an approximation of a uniform distribution of 5 bins (without contiguity constraint) after which the bins are subsampled to have the same amount of sentences as the smallest bin. Details about labels and frequency of each bin are in \autoref{tab:ArgmaxCount}.

\begin{table}[!htbp]\centering
\footnotesize
\captionsetup{font=footnotesize}
\begin{tabular}{lp{0.5\linewidth}rr}\toprule
\multicolumn{3}{c}{\textbf{ArgmaxCount}} \\\cmidrule{1-3}
\textbf{Bin} &\textbf{Labels in bin} &\textbf{Bin Size} \\\midrule
1 &[`b', `c', `d', `f', `g', `h', `k', `l', `m', `p', `r', `t', `u', `w', `y', `z'] &9413 \\\midrule
2 &[`n', `o', `s'] &9490 \\\midrule
3 &[`i'] &9816 \\\midrule
4 &[`a'] &11784 \\\midrule
5 &[`e'] &9900 \\\midrule
&\textbf{Total} &50,403 \\
\bottomrule
\end{tabular}
\caption{Bin sizes, labels in bin and total data size for \task{ArgmaxCount}. The labels correspond to the character with the maximum frequency in an example. The bin for `e' has been subsampled from 59,497 to 9900 to ensure a relatively uniform distribution across bins.}\label{tab:ArgmaxCount}
\end{table}

\newpage
\newpage
\section{Model Parameters}\label{sec:model_params}
The models used in this study along with their parameter sizes are in \autoref{tab:models}.

\begin{table}[!htp]\centering
 \footnotesize
\scriptsize
\begin{tabular}{lrrrrrr}\toprule
\multicolumn{6}{c}{\textbf{Models}} \\\cmidrule{1-6}
\textbf{Name} &\textbf{Enc-Dec} &\textbf{Hid} &\textbf{MLP} &\textbf{Att} &\textbf{|$\theta$|} \\\midrule
\bert &12-0 &768 &3072 &12 &110M \\
\vitmae &12-8 &768 &3072 &12 &86M \\\midrule
\pixel-base &12-8 &768 &3072 &12 &86M \\
\pixel-bigrams &12-8 &768 &3072 &12 &86M \\
\pixel-small &12-4 &384 &1536 &6 &22M \\
\pixel-small-bigrams &12-4 &384 &1536 &6 &22M \\
\pixel-small-words &12-4 &384 &1536 &6 &22M \\
\bottomrule
\end{tabular}
\caption{Size of the probed models.}
\label{tab:models}
\end{table}

\newpage
\section{\pixel vs \pixel-bigrams}\label{pixel-v-bigrams}

\begin{figure}[!h]
    \captionsetup{font=footnotesize}
    \centering
    \scriptsize
    \includegraphics[scale=0.75]{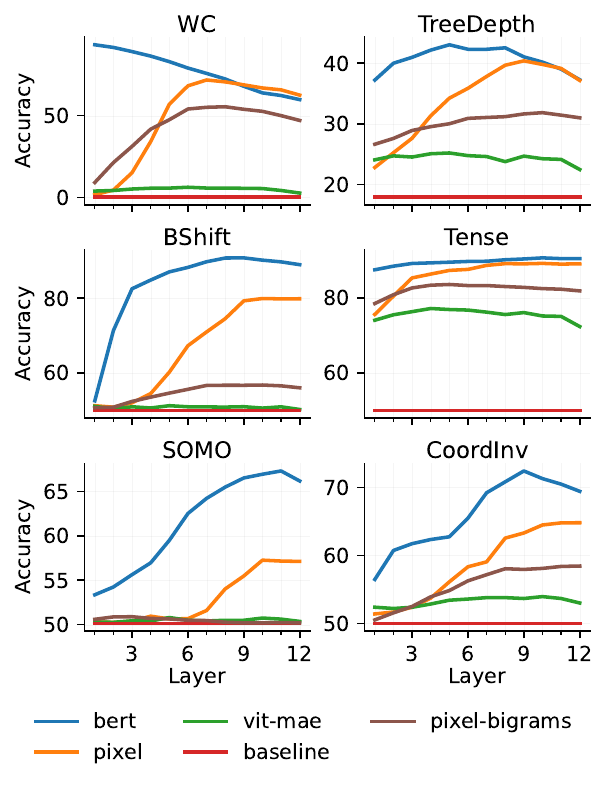}
    
    \caption{Selected linguistic probing results for \pixel (orange), \pixel-bigrams (brown), \bert (blue) and \vitmae (green).}
    \label{fig:base-models}
\end{figure}

Select probing results for \pixel and \pixel-bigrams base models are in \autoref{fig:base-models}. As observed, \pixel-bigrams performs worse than \pixel across all probing tasks. We theorize that even though bigrams rendering imposes some structure on the input text, it results in a loss of word boundary information and longer sequences. The rendering strategy adds extra space even within words to ensure that one patch has only two characters, and creates more ambiguity about the structure of the word. 
This is most prominently seen in the tasks that test for word level information within a sentence - namely, \texttt{WC}, \texttt{BShift} and \texttt{SOMO}. For the later 2, \pixel-bigrams barely outperforms the majority baseline.

\end{document}